%% file: main.tex
\documentclass[11pt,a4paper]{article}
\usepackage[hyperref]{emnlp2020}
\usepackage{times}
\usepackage{latexsym}

\usepackage{boxedminipage} % for the examples in figures
\usepackage{graphicx}
\usepackage{multirow}
\usepackage{amsmath,amssymb,amsthm,bbm,mathtools,bm}
\usepackage{algorithm,algorithmicx,algpseudocode}

\definecolor{gred}{RGB}{219,68,55}
\definecolor{gblue}{RGB}{66,133,244}
\definecolor{gyellow}{RGB}{244,180,0}
\definecolor{ggreen}{RGB}{15,157,88}
\definecolor{ggrey}{RGB}{115,115,115}
\newcommand{\error}[1]{\textcolor{gred}{\textbf{#1}}} % highlight errors in examples
\newcommand{\reph}[1]{\textcolor{ggreen}{\textbf{#1}}} % highlight repetition in first page example
\usepackage{microtype}

\aclfinalcopy % Uncomment this line for the final submission
\newcommand{\norm}[1]{\left\lVert#1\right\rVert}

\title{Bootstrapped Q-learning with Context Relevant Observation\\ Pruning to Generalize in Text-based Games}

\author{Subhajit Chaudhury \\
	IBM Research AI \\
	\texttt{subhajit@jp.ibm.com} \\\And
	Daiki Kimura \\
	IBM Research AI \\
	\texttt{daiki@jp.ibm.com} \\\And
	Kartik Talamadupula \\
	IBM Research AI \\
	\texttt{krtalamad@us.ibm.com}
	\AND
	Michiaki Tatsubori \\
	IBM Research AI \\
	\texttt{mich@jp.ibm.com} \\\And
	Asim Munawar \\
	IBM Research AI \\
	\texttt{asim@jp.ibm.com} \\\And
	Ryuki Tachibana \\
	IBM Research AI \\
	\texttt{ryuki@jp.ibm.com} }

\date{}

\begin{document}
	\maketitle
	\begin{abstract}
		We show that Reinforcement Learning~(RL) methods for solving Text-Based Games~(TBGs) often fail to generalize on unseen games, especially in small data regimes. To address this issue, we propose \textbf{C}ontext \textbf{R}elevant \textbf{E}pisodic \textbf{S}tate \textbf{T}runcation~(\textbf{CREST}) for irrelevant token removal in observation text for improved generalization. Our method first trains a base model using Q-learning, which typically overfits the training games. The base model's action token distribution is used to perform observation pruning that removes irrelevant tokens. A second bootstrapped model is then retrained on the pruned observation text. Our bootstrapped agent shows improved generalization in solving unseen TextWorld games, using $10$x-$20$x fewer training games compared to previous state-of-the-art~(SOTA) methods despite requiring less number of training episodes. 
		
	\end{abstract}

	\vspace{-0.3cm}

	\section{Introduction}

	Reinforcement Learning~(RL) methods are increasingly being used for solving sequential decision-making problems from natural language inputs, like text-based games~\cite{narasimhan2015language, he2016deep, yuan2018counting, zahavy2018learn} chat-bots~\cite{serban2017deep} and personal conversation assistants~\cite{dhingra2017towards, li2017end, wu2016google}. In this work, we focus on Text-Based Games~(TBGs), which require solving goals like \textit{``Obtain coin from the kitchen''}, based on a natural language description of the agent's observation of the environment. To interact with the environment, the agent issues text-based action commands~(``\textit{go west}'') upon which it receives a reward signal used for training the RL agent.
	
	\input{first_page_figure}

	\begin{figure*}[tb]
		\centering
			\includegraphics[width=0.9\textwidth]{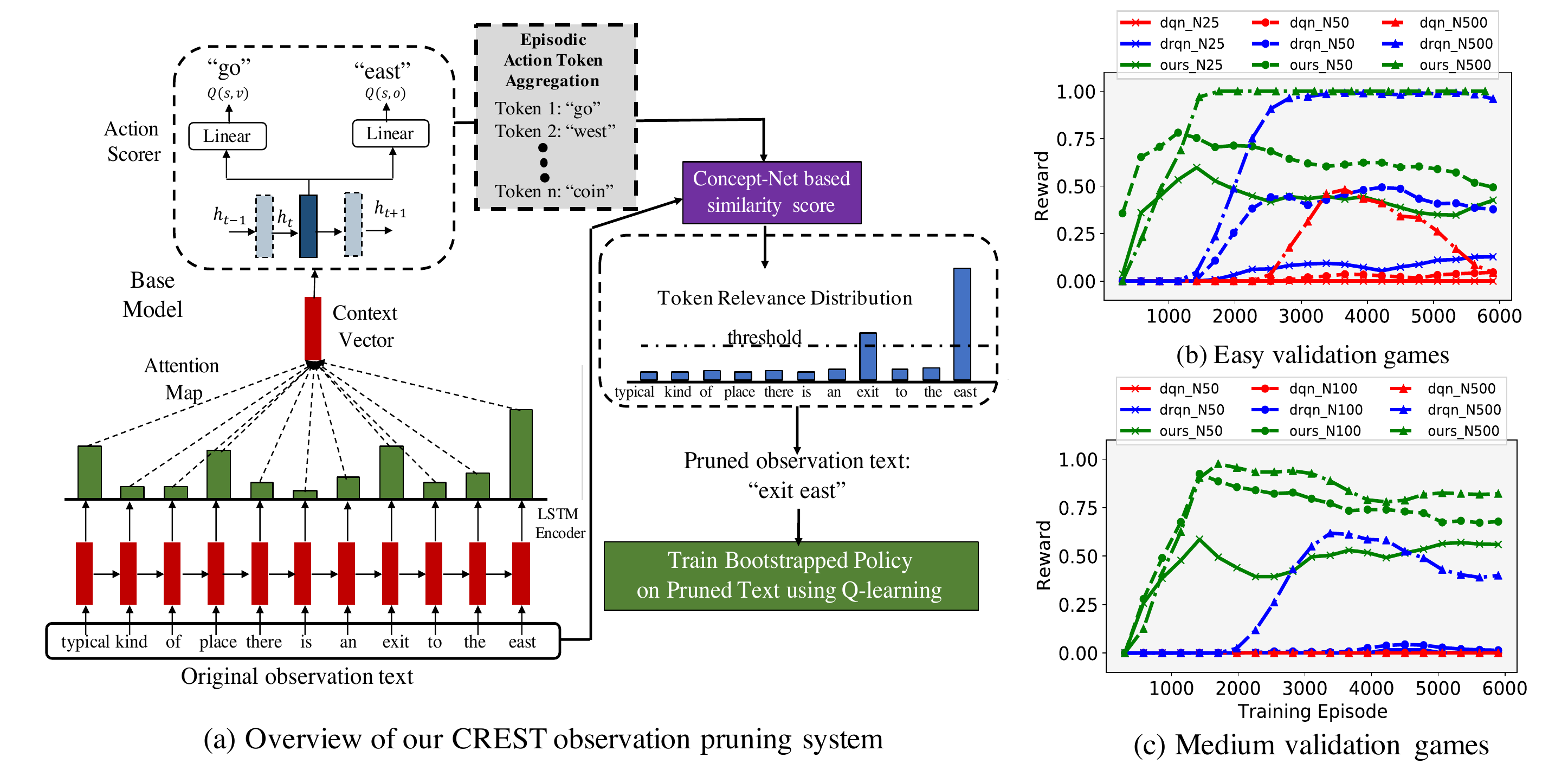}
		\caption{(a) Overview of Context Relevant Episodic State Truncation~(CREST) module using Token Relevance Distribution for observation pruning. Our method shows better generalization from $10$x-$20$x less number of training games and faster learning with fewer episodes on (b) ``easy'' and (c) ``medium'' validation games.}
		\label{fig:model}
	\end{figure*}

	% generalization problem
	Traditional text-based RL methods focus on the problems of partial observability and large action spaces. However, the topic of generalization to unseen TBGs is less explored in the literature.  We show that previous RL methods for TBGs often show poor generalization to unseen test games. We hypothesize that such overfitting is caused due to the presence of irrelevant tokens in the observation text, which might lead to action memorization.
	% ~(eg. every time agent. 
	To alleviate this problem, we propose \textbf{CREST}, which first trains an overfitted \textit{base} model on the original observation text in training games using Q-learning. Subsequently, we apply observation pruning such that, for each episode of the training games, we remove the observation tokens that are not semantically related to the base policy's action tokens. Finally, we re-train a bootstrapped policy on the pruned observation text using Q-learning that improves generalization by removing irrelevant tokens. Figure~\ref{fig_firstpage} shows an illustrative example of our method. Experimental results on TextWorld games~\cite{cote2018textworld} show that our proposed method generalizes to unseen games using almost $10$x-$20$x fewer training games compared to SOTA methods; and features significantly faster learning.

	\vspace{-0.2cm}
	\section{Related Work} %% Please check here!!
	\vspace{-0.2cm}
	LSTM-DQN~\cite{narasimhan2015language} is the first work on text-based RL combining natural language representation learning and deep Q-learning.
	LSTM-DRQN~\cite{yuan2018counting} is the state-of-the-art on TextWorld CoinCollector games, and addresses the issue of partial observability by using memory units in the action scorer. 
	\citet{fulda2017can} proposed a method for affordance extraction via word embeddings trained on a Wikipedia corpus. 
	AE-DQN~(Action-Elimination DQN) -- which is a combination of a Deep RL algorithm with an action eliminating network for sub-optimal actions -- was proposed by Zahavy et al.~\cite{zahavy2018learn}. 
	Recent methods~\cite{adolphs2019ledeepchef, ammanabrolu2018playing, ammanabrolu2020graph,yin2019learn, adhikari2020learning} use various heuristics to learn better state representations for efficiently solving complex TBGs.
	
	\begin{table*}[t]
		%\scriptsize
		\centering
		\caption{ The average success rate of various methods on 20 unseen test games. Experiments were repeated on 3 random seeds. Our method trained on almost $20x$ fewer data has a similar success rate to state-of-the-art methods.}
		
		\begin{tabular}{|c|ccc|ccc|cc|}
			\hline
			\multirow{2}{*}{Methods} & \multicolumn{3}{c|}{Easy} & \multicolumn{3}{c|}{Medium} & \multicolumn{2}{c|}{Hard} \\ \cline{2-9} 
			& N25 & N50 & N500 & N50 & N100 & N500 & N50 & N100 \\ \hline
			LSTM-DQN~(no att) & 0.0 & 0.03 & 0.33 & 0.0 & 0.0 & 0.0 & 0.0 & 0.0 \\
			LSTM-DRQN~(no att) & 0.17 & 0.53 & 0.87 & 0.02 & 0.0 & 0.25 & 0.0 & 0.0 \\
			LSTM-DQN~(+attn) & 0.0 & 0.03 & 0.58 & 0.0 & 0.0 & 0.0 & 0.0 & 0.0 \\
			LSTM-DRQN~(+attn) & 0.32 & 0.47 & 0.87 & 0.02 & 0.06 & 0.82 & 0.02 & 0.08 \\ \hline
			\textbf{Ours~(ConceptNet+no att)} & 0.47 & 0.5 & 0.98 & \textbf{0.75} & 0.67 & \textbf{0.97} & 0.62 & \textbf{0.92} \\
			\textbf{Ours~(Word2vec+att) }& 0.67 & 0.82 & \textbf{1.0} & 0.57 & 0.92 & 0.95 & 0.77 & \textbf{0.92} \\
			\textbf{Ours~(Glove+att)} & 0.70 & \textbf{0.97} & \textbf{1.0} & 0.67 & 0.72 & 0.90 & 0.1 & 0.63 \\ 
			\textbf{Ours~(ConceptNet+att)} &\textbf{ 0.82} & 0.93 & \textbf{1.0} & 0.67 & \textbf{0.95} & \textbf{0.97} & \textbf{0.93} & 0.88 \\ \hline
		\end{tabular}
	\end{table*}

	\section{Our Method}

	\vspace{-0.2cm}
	\subsection{Base model}
	\label{sec:base}
	We consider the standard sequential decision-making setting: a finite horizon Partially Observable Markov Decision Process (POMDP), represented as $(s, a, r, s')$, where $s$ is the current state, $s'$ the next state, $a$ the current action, and $r(s, a)$ is the reward function. The agent receives state description $\bm{s}_t$ that is a combination of text describing the agent's observation and the goal statement. The action consists of a combination of verb and object output, such as ``go north'', ``take coin'', etc. The overall model has two modules: a representation generator, and an action scorer as shown in Figure~\ref{fig:model}. The observation tokens are fed to the embedding layer, which produces a sequence of vectors $\bm{x}^t = \{x^t_1, x^t_2,..., x^t_{N_t}\}$, where $N_t$ is the number of tokens in the observation text for time-step $t$. We obtain hidden representations of the input embedding vectors using an LSTM model as $h^t_i = f(x^t_i, h^t_{i-1})$. We compute a context vector~\cite{bahdanau2014neural} using attention on the $j^{th}$ input token as,
	
	\vspace{-0.4cm}
	\begin{align}
	e^t_{j} = v^T\tanh(W_hh^t_j + b_{attn})\\
	\alpha^t_{j} = \text{softmax}(e^t_{j})
	\end{align}
	\noindent where $W_h$, $v$ and $b_{attn}$ are learnable parameters. The context vector at time-step $t$ is computed as the weighted sum of embedding vectors as $c^t = \sum_{j=1}^{N_t}{\alpha^t_{j}h^t_j}$. The context vector is fed into the action scorer, where two multi-layer perceptrons (MLPs), $Q(s, v)$ and $Q(s, o)$ produce the Q-values over available verbs and objects from a shared MLP's output. The original works of \citet{narasimhan2015language,yuan2018counting} do not use the attention layer. LSTM-DRQN replaces the shared MLP with an LSTM layer, so that the model remembers previous states, thus addressing the partial observability in these environments.

	Q-learning~\cite{watkins1992q,mnih2015human} is used to train the agent. %'s optimal Q-value functions. 
	The parameters of the model are updated, by optimizing the following loss function obtained from the Bellman equation~\cite{sutton1998introduction},
	\begin{equation}
	\mathcal{L} = \norm{Q(s,a) - \mathbb{E}_{s,a}\bigg[ r + \gamma\max_{a'}Q(s',a')\bigg]}_2\\
	\end{equation}
	\noindent where $Q(s, a)$ is obtained as the average of verb and object Q-values, $\gamma \in (0, 1)$ is the discount factor. The agent is given a reward of $1$ from the environment on completing the objective. We also use episodic discovery bonus ~\cite{yuan2018counting} as a reward during training that introduces curiosity~\cite{pathak2017curiosity} encouraging the agent to uncover unseen states for accelerated convergence.
	
	\subsection{Context Relevant Episodic State Truncation (CREST)}
	\label{sect:crest}
	
	\begin{figure}[tb]
		\centering
		\includegraphics[width=1.0\linewidth]{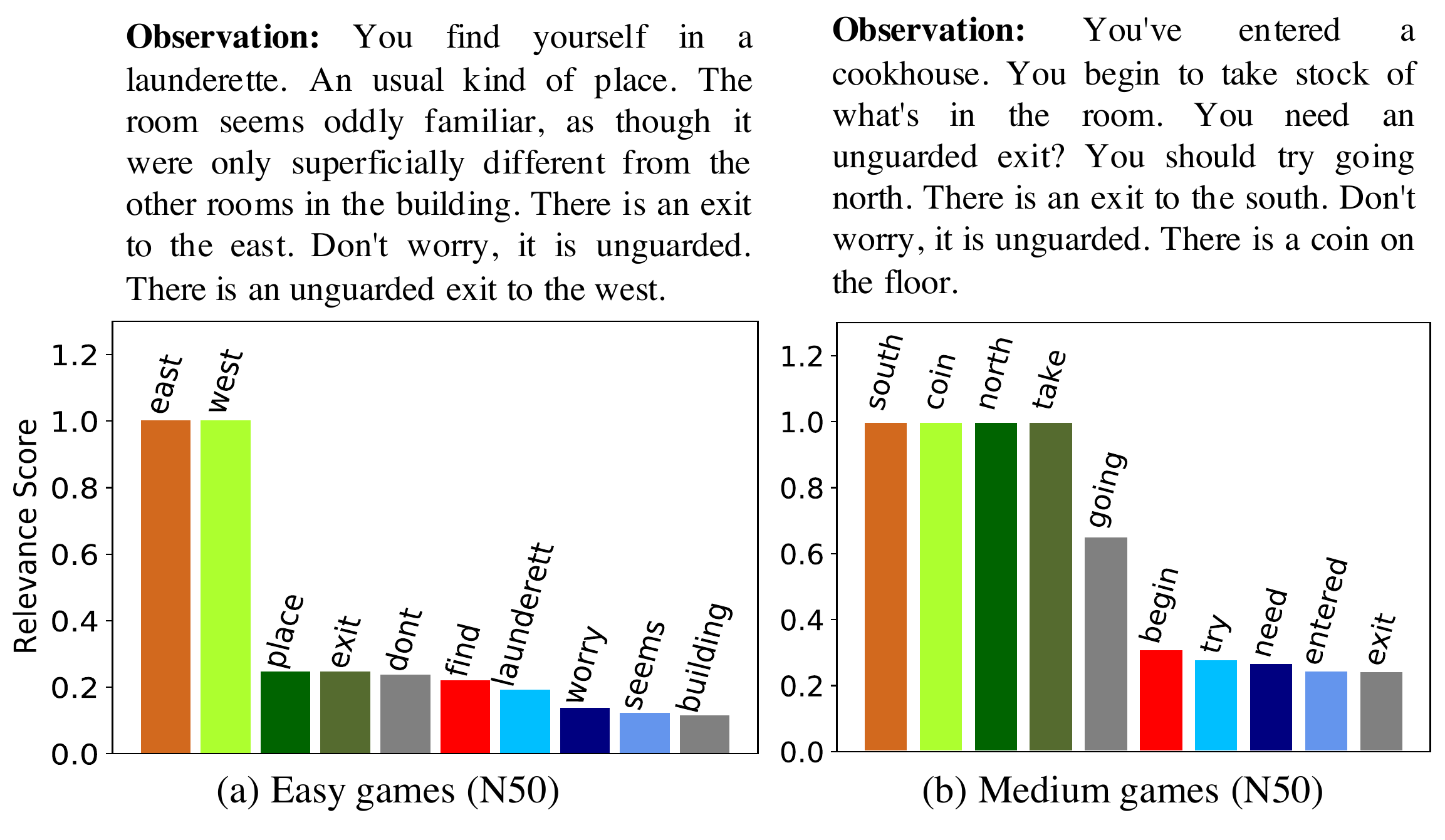}
		\caption{Ranking of context-relevant tokens from observation text by our token relevance distribution.}
		\label{fig:relevance}
	\end{figure}

	Traditional LSTM-DQN and LSTM-DRQN methods, trained on observation text containing irrelevant textual artifacts~(like ``You don't like doors?'' in Figure~\ref{fig_firstpage}), that leads to overfitting in small data regimes. Our CREST module removes unwanted tokens in the observation that do not contribute to decision making. Since the base policy overfits on the training games, the action commands issued by it can successfully solve the training games, thus yielding correct (observation text, action command) pairs for each step in the training games. Therefore, by only retaining tokens in the observation text that are contextually similar to the base model's action command, we remove unwanted tokens in the observation, which might otherwise cause overfitting. Figure~\ref{fig:model}(a) shows the overview of our method.
	
	\begin{figure*}[t]
		\centering
		\includegraphics[width=0.9\linewidth]{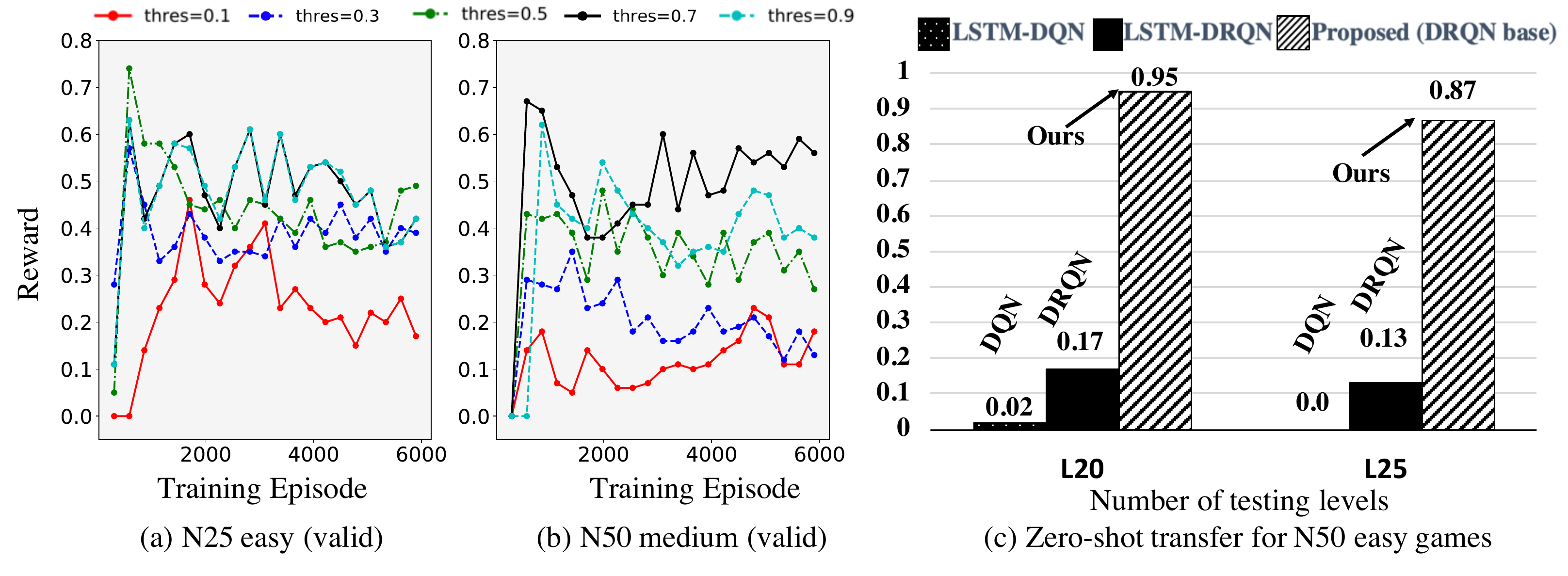}
		\caption{Comparison of validation performance for various thresholds on (a) \textit{easy} and (b) \textit{medium} games, (c) Our method trained on $L15$ games and tested on $L20$ and $L25$ games significantly outperforms the previous methods.}
		\label{fig:zeroshot}
	\end{figure*}

	We use three embeddings to obtain token relevance: (1) Word2Vec~\cite{mikolov2013distributed}; (2) Glove~\cite{pennington2014glove}; and (3) Concept-net~\cite{liu2004conceptnet}. 

	The distance between tokens is computed using cosine similarity, $D(\bm{a},\bm{b})$.

	\noindent \textbf{Token Relevance Distribution (\textbf{TRD})}:
	We run inference on the overfitted base model for each training game (indexed by $k$) and aggregate all the action tokens issued for that particular game as the Episodic Action Token Aggregation~(EATA), $\mathcal{A}^k$. For each token $w_i$ in a given observation text $\bm{o}_t^k$ at step $t$ for the $k^{th}$ game, we compute the Token Relevance Distribution~(TRD) $\mathcal{C}$ as:
	\vspace{-0.1cm}
	\begin{equation}
	\mathcal{C}(w_i, \mathcal{A}^k) = \max_{a_j \in \mathcal{A}^k} D(w_i, a_j)\;\;\forall\; w_i \in \bm{o}_t^k ,
	\end{equation}
	\vspace{-0.1cm}

	\noindent where the $i^{th}$ token $w_i$'s score is computed as the maximum similarity to all tokens in $\mathcal{A}^k$. This relevance score is used to prune irrelevant tokens in the observation text by creating a hard attention mask using a threshold value. Figure~\ref{fig:relevance} presents examples of TRD's from observations highlighting which tokens are relevant for the next action. Examples of token relevance are shown in the appendix.

	\noindent \textbf{Bootstrapped model}: The bootstrapped model is trained on the pruned observation text by removing irrelevant tokens using TRDs. Same model architecture and training methods as the base model are used. During testing, TRDs on unseen games are computed as $\mathcal{C}(w_i, \mathcal{G})$, by global aggregation of action tokens, $\mathcal{G} = \bigcup_k \mathcal{A}^k$, that combines the EATA for all training games. This approach retains all relevant action tokens to obtain the training domain information during inference assuming similar domain distribution between training and test games.

	\vspace{-0.1cm}
	\section{Experimental Results}
	\label{sec:results}
	\vspace{-0.1cm}
	\textbf{Setup:}
	We used \textit{easy}, \textit{medium}, and \textit{hard} modes of the Coin-collector Textworld~\cite{cote2018textworld, yuan2018counting} framework for evaluating our model's generalization ability. The agent has to collect a \textit{coin} that is located in a particular room. 
	
	We trained each method on various numbers of training games~(denoted by N\#) to evaluate generalization ability from a few number of games.
	
	\noindent \textbf{Quantitative comparison:}
	We compare the performance of our proposed model with LSTM-DQN~\cite{narasimhan2015language} and LSTM-DRQN~\cite{yuan2018counting}. 

	Figure~\ref{fig:model}(b) and \ref{fig:model}(c) show the reward of various trained models, with increasing training episodes on \textit{easy} and \textit{medium} games. Our method shows improved out-of-sample generalization on validation games with about $10$x-$20$x fewer training games ($500$ vs. $25, 50$) with accelerated training using drastically fewer training episodes compared to previous methods.

	We report performance on unseen test games in Table 1. Parameters corresponding to the best validation score are used. Our method trained with $N25$ and $N50$ games for \textit{easy} and \textit{medium} levels respectively achieves performance similar to $500$ games for SOTA methods. We perform ablation study with and without attention in the policy network and show that the attention mechanism alone does not substantially improve generalization. We also compare the performance of various word embeddings for TRD computation and find that ConceptNet gives the best generalization performance.

	\noindent \textbf{Pruning threshold:} In this experiment, we test our method's response to changing threshold values for observation pruning. Figure~\ref{fig:zeroshot}(a) and Figure~\ref{fig:zeroshot}(b) reveals that thresholds of $0.5$ for easy games and $0.7$ for medium games, gives the best validation performance. A very high threshold might remove relevant tokens also, leading to failure in the training; whereas a low threshold value would retain most irrelevant tokens, leading to over-fitting.

	\noindent \textbf{Zero-shot transfer}: In this experiment, agents trained on games with quest lengths of $15$ rooms were tested on unseen game configurations with quest length of $20$ and $25$ rooms respectively without retraining, to study the zero-shot transferability of our learned agents to unseen configurations. The results in the bar charts of Figure~\ref{fig:zeroshot}(c) for $N50$ easy games show that our proposed method can generalize to unseen game configurations significantly better than previous state-of-the-art methods on the coin-collector game.

	\noindent \textbf{Generalizability to other games}: 
	In the above experimental section, we reported the results on the coin-collector environment, where the noun and verbs used in the training and testing games have a strong overlap. In this section, we present some discussion about the generalizability of this method to other games, where the context-relevant tokens for a particular game might be never seen in the training games.
	
	To test the generalizability of our method, we performed experiments on a different type of a game~(cooking games) considered in \cite{adolphs2019ledeepchef}. An example observation looks like this: ``
	\textsc{\ldots You see a fridge. The fridge contains some water, a diced cilantro and a diced parsley. You wonder idly who left that here. Were you looking for an oven? Because look over there, it's an oven. Were you looking for a table? Because look over there, it's a table. The table is massive. On the table you make out a cookbook and a knife. You see a counter. However, the counter, like an empty counter, has nothing on it \ldots}''. The objective of this game is to prepare a meal following the recipe found in the kitchen and eat it.
	
	We took 20 training and 20 testing cooking games with unseen items in test observations. Training action commands are obtained from an oracle. From the training games, we obtain noun action tokens as, {'onion', 'potato', 'parsley', 'apple', 'counter', 'pepper', 'meal', 'water', 'fridge', 'carrot'}. Using our token relevance method (using concept-net embeddings) described in Section 3.2, we obtain scores for unseen cooking related nouns during test as, {"banana": 0.45, "cheese": 0.48, "chop": 0.39, "cilantro": 0.71, "cookbook": 0.30, "knife": 0.13, "oven": 0.52, , "stove": 0.48, "table": 0.43}. 
	
	Although these nouns were absent in the training action distribution, our proposed method can assign a high score to these words (except “knife”), since they are similar in concept to training actions. An appropriate threshold (eg.,th=0.4) can retain most tokens, which can be automatically tuned using validation games as shown in Fig 4(a) and 4(b) in the paper. Thus, as described in Section 5, assuming some level of overlap between training and testing knowledge domains, our method is generalizable and can reduce overfitting for RL in NLP. Large training and testing distribution gap is a much more difficult problem, even for supervised ML and conventional RL settings, and is out of the scope of this paper.

	\vspace{-0.3cm}
	\section{Conclusion}
	\label{sec:supplemental}
	\vspace{-0.3cm}
	We present a method for improving generalization in TBGs using irrelevant token removal from observation texts. Our bootstrapped model trained on the salient observation tokens obtains generalization performance similar to SOTA methods, with $10$x-$20$x fewer training games, due to better generalization; and shows accelerated convergence. 
	In this paper, we have restricted our analysis to TBGs that feature similar domain distributions in training and test games. In the future, we wish to handle the topic of generalization in the presence of domain differences such as novel objects, and goal statements in test games that were not seen in training.

	\bibliography{anthology,acl2020,bibs/summarization}
	\bibliographystyle{acl_natbib}

	\appendix
	
	\begin{figure*}[tb]
		\centering
		\includegraphics[width=0.88\textwidth]{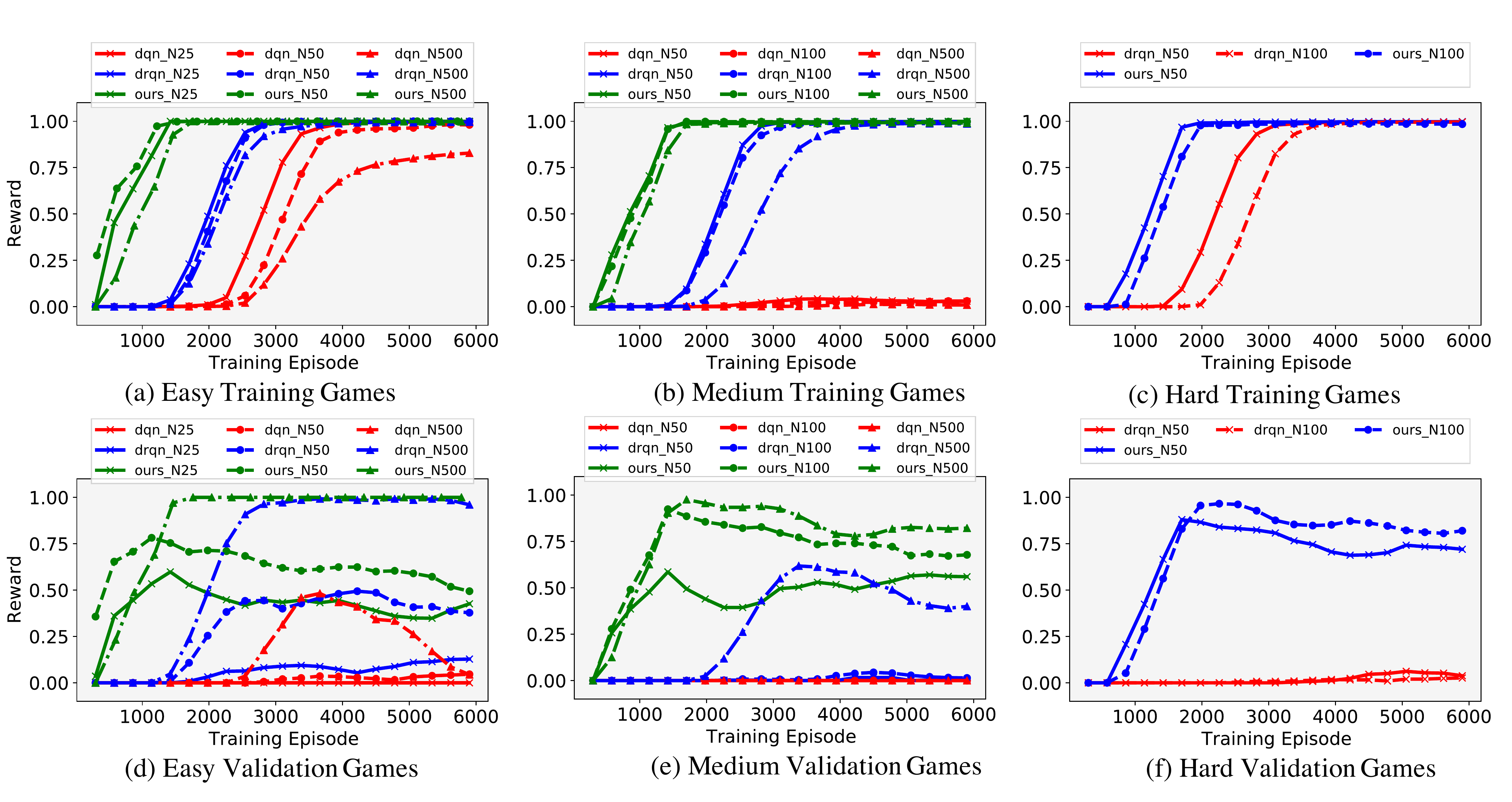}
		\caption{Training and validation games learning curve for various games. The metric of measurement (y-axis) is the avergare of final reward of 1.0 on completion of the quest and 0.0 otherwise, thus measuring the average success rate. Our method shows better generalization from significantly less number of training games and faster learning with fewer episodes for all cases of ``easy'', ``Medium'' and ``hard'' validation games.}
		\label{fig:model_appendix}
	\end{figure*}
	
	\section{Description of Text-based games}
	
	We used Textworld~\cite{cote2018textworld} framework for evaluating our model's generalization ability on text-based games. For each game, the agent is provided with a goal statement and an observation text describing the current state of the world around it only. The agent has to overcome partial observability using memory because it never sees the full state of the world. The games are inspired by the chain experiments used in \cite{plappert2017parameter, osband2016deep} for evaluating exploration in RL policies. The agent has to navigate through various rooms that are randomly connected to form a chain, finally reaching the goal state. We ensure that the goal statement does not contain navigational instructions by using the ``\texttt{--only-last-action}'' option. The agent is rewarded only when it successfully achieves the end-goal. We use a quest length~(number of rooms to travel before reaching goal state) of $L15$ for training our policies. 
	
	We use the Coin-collector environment for evaluating our experiments, where the agent has to collect a \textit{coin} that is located in a particular room. For different games the location of coin and interconnectivity between the rooms are different. We experiment with \textbf{three modes} of this challenge according to ~\cite{yuan2018counting}: \textit{easy}, there are no distractor rooms (dead-ends) along the optimal path, and the agent needs to choose a command that only depends on the previous state; \textit{medium}, there is one distractor per room in the optimal path and the agent has to issue a reverse command of its previous command to come out of such distractor rooms; \textit{hard}, there are two distractors per rooms in the optimal path and the agent has to issue a reverse command of its previous command to come out of such distractor rooms, in addition to remembering longer into the past to successfully keep track of which paths it has already traveled.
	\vspace{-0.2cm}
	\section{Experimental Setup}
	\vspace{-0.2cm}
	
	For the Textworld coin collector environment, we use 10 verbs and 10 nouns in the vocabulary which is learned using Q learning. This is different from previous methods that use only 2 words and 5 objects thus increasing the complexity of Q learning slightly. However, it is to be noted that the problem of generalization exists even for less number of action tokens as reported in \cite{yuan2018counting} and is not significantly aggravated on a slightly larger action space (10 vs 100 combinations). The configuration used in our base and bootstrapped model learning is the same as the previous methods with the only change being the addition of the attention layer. We trained each environment for 6000 epochs with annealing of 3600 epochs from a starting value of 1.0 to 0.2. Each training experiment took about $5$-$6$ hours for completion. Our experiments were conducted on a Ubuntu16.04 system with a Titan X (Pascal) GPU. We use a single LSTM network with 100 dimensional hidden units in the representation generator. For the action scorer, a single LSTM network with 64-dim hidden unit (for DRQN ) and two MLPs for verb and object Q-values were used. The number of trainable parameters in our policy network is 128,628 for the model with attention and 125,364 without attention.

	\begin{figure*}[tb]
		\centering
		\includegraphics[width=0.88\textwidth]{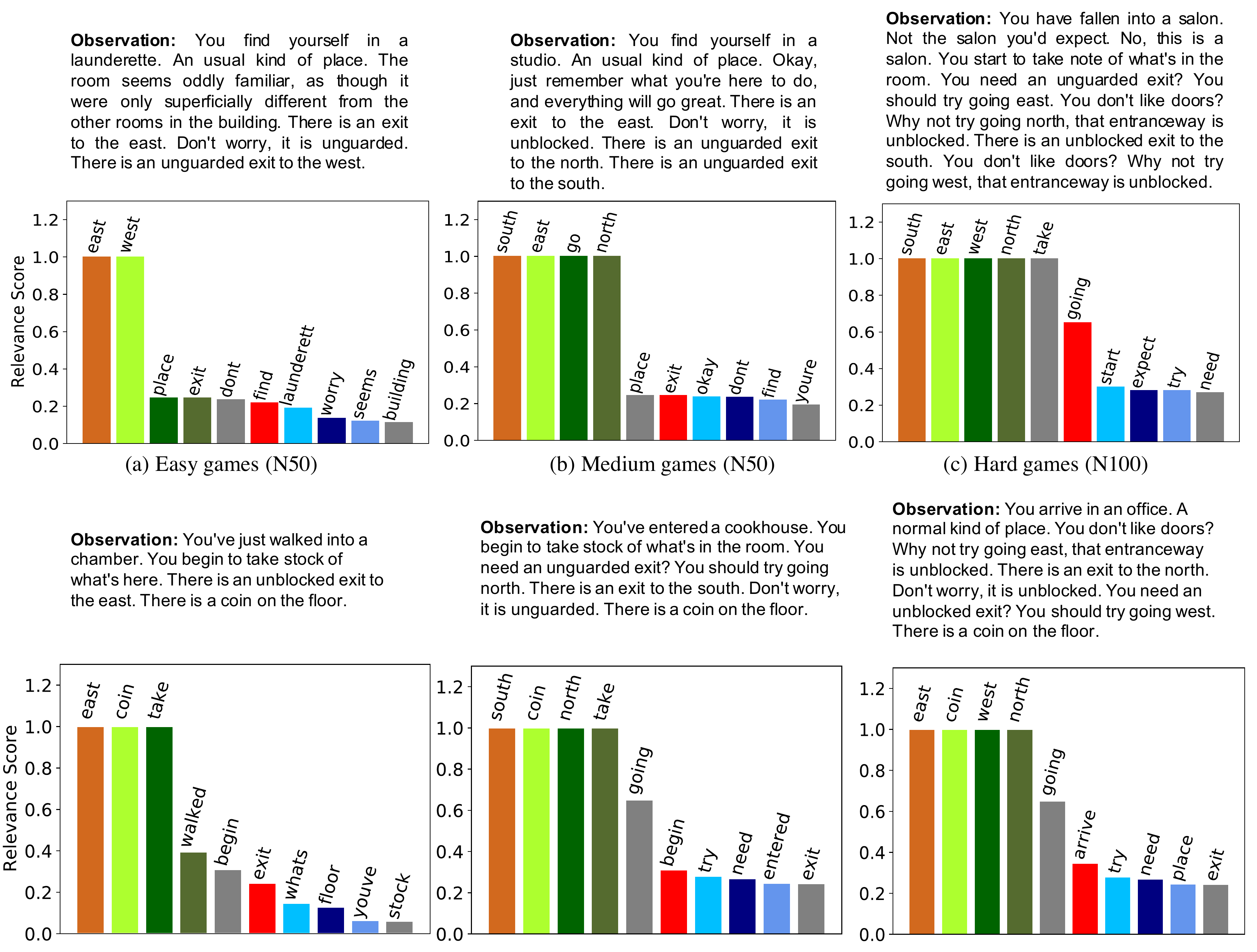}
		\caption{Showing the relevance distribution of observation token for easy, medium and hard games along with original observation text. The top row shows a non-terminal observation where the ``coin'' is not present. The second row shows terminal states. Each relevance score is bounded between [0,1]. The bootstrapped model is trained on tokens that have relevance above some threshold to remove irrelevant tokens.}
		\label{fig:relevance_appendix}
	\end{figure*}
	
	\begin{table*}[tb]
		\centering
		
		\begin{tabular}{|c|ccc|ccc|cc|}
			\hline
			\multirow{2}{*}{Methods} & \multicolumn{3}{c|}{Easy} & \multicolumn{3}{c|}{Medium} & \multicolumn{2}{c|}{Hard} \\ \cline{2-9} 
			& N25 & N50 & N500 & N50 & N100 & N500 & N50 & N100 \\ \hline
			\multicolumn{9}{|c|}{Evaluate on L20} \\ \hline
			LSTM-DQN~(+attn) & 0.0 & 0.02 & 0.58 & 0.0 & 0.0 & 0.0 & 0.0 & 0.0 \\
			LSTM-DRQN~(+attn) & 0.25 & 0.17 & 0.90 & 0.0 & 0.02 & 0.53 & 0.0 & 0.02 \\
			Ours~(ConceptNet+att) & \textbf{0.65} &\textbf{ 0.95 }& \textbf{1.0} & \textbf{0.67} & \textbf{0.82} & \textbf{0.98} & \textbf{0.98} & \textbf{0.80} \\ \hline
			\multicolumn{9}{|c|}{Evaluate on L25} \\ \hline
			LSTM-DQN~(+attn) & 0.0 & 0.0 & 0.48 & 0.0 & 0.0 & 0.00 & 0.0 & 0.0 \\
			LSTM-DRQN~(+attn) & 0.07 & 0.13 & 0.88 & 0.0 & 0.02 & 0.50 & 0.02 & 0.0 \\
			Ours~(ConceptNet+att) & \textbf{0.65} & \textbf{0.87} & \textbf{1.0} & \textbf{0.52 }& \textbf{0.77} & \textbf{0.92} & \textbf{0.95} & \textbf{0.88} \\ \hline
		\end{tabular}
		\caption{Average succes in zero-shot transfer to other configurations. We trained the RL policies for $L15$ games and test the performance on $L20$ and $L25$ unseen game configurations. CREST significantly outperforms the previous methods on such tasks for all cases of \textit{easy}, \textit{medium} and \textit{hard} games. }
		\label{table:zero}
	\end{table*}
	
	In our experiments, we wish to investigate the generalization property of our method on a small number of training games. To that end, we wish to answer the following questions: (1) Can our proposed observation masking system out-perform previous RL methods for TBGs using less training data with accelerated learning?, (2) Is there a positive correlation between the strength of observation masking and generalization performance? and (3) Does our method understand the semantic meaning of the games to perform zero-shot generalization to unseen configurations of TBGs?
	
	\section{Improved generalization by CREST}
	The generalization ability of the learned base policy is measured by the performance on unseen games that were not used during training the policy network. We measure the reward obtained by the agent in each episode which is the metric of success in our experiments. During the evaluation of unseen games, only the environment reward is used and the episodic discovery bonus is turned off. Since a reward of 1.0 is obtained on the completion of the game, the average reward can also be interpreted as the average success rate in solving the games.
	
	The verb and object tokens corresponding to the maximum Q-value are chosen as the action command. Traditional LSTM-DQN and LSTM-DRQN methods are trained on observation text descriptions that include irrelevant textual artifacts, which might lead to overfitting in small data regimes.  To demonstrate this effect, we plot the performance of LSTM-DRQN (SOTA on coin-collector) and LSTM-DQN on the Coin-Collector \textit{easy}, \textit{medium}, and \textit{hard} games on various training and $20$ unseen validation games in Figure~\ref{fig:model_appendix}. In each training episode, a random batch of games is sampled from the available training games and Q-learning is performed. 
	
	While for a large number of training games~(500), the SOTA policies can solve most of the validation games (especially for easy games). However, the performance degrades significantly for less number of training games. On the other hand, the training performance shows a 100\% success rate indicating overfitting. This kind of behavior might occur if the agent associates certain action commands to irrelevant tokens in the observation. For example, the agent might encounter games in training where observation tokens, ``a typical kind of place'' correspond to the action of ``go east''. In this case, the agent might learn to associate such irrelevant tokens to the ``go east'' command without actually learning the true dependency on tokens like ``there is a door to the east''.
	
	\subsection{Quantitative evaluation of generalization}
	Our proposed method shows better generalization performance as is evident from Figure~\ref{fig:model_appendix}. Both training and validation performance increases with increasing training episodes, indicating good generalization. Slight overfitting is evidenced if the agent is trained for a longer duration. The policy parameters corresponding to the best validation score is used for evaluating unseen test games. Our policy shows better performance due to training on only context-relevant tokens after the removal of unwanted tokens from observation text. We show the visualization of Token Relevance Distributions~(TRDs) obtained by our method for \textit{easy},  \textit{medium}, and \textit{hard} games in Figure~\ref{fig:relevance_appendix}. Each token has a similarity score between 0 and 1, indicating how relevant it is for making decisions about the next action. Tokens with a score less than a threshold are pruned. We also perform such observation pruning in the testing phase. Therefore, our proposed method learns on such clean observation texts which are also tested on unseen pruned texts, which leads to improved generalization.
	
	\vspace{-0.2cm}
	\subsection{Zero-shot transfer}
	While in the previous experiments the training and evaluation games had the same quest length configuration, in this experiment we evaluate our method on games with different configurations of coin-collector never seen during training. Specifically, during training, we use games with quest lengths of $15$ rooms. The models trained on such configuration are tested on games with quest length of $20$ and $25$ rooms respectively without any retraining. This is aimed to study the zero-shot transferability to other configurations that the agent has never encountered before. The results are shown in Table~\ref{table:zero} for all modes of the coin-collector games, show that our proposed observation masking method can also generalize to unseen game configurations with increased quest length and largely outperforms the previous state-of-the-art methods. Our method, \textbf{CREST} learns to retain important tokens in the observation text which leads to a better semantic understanding resulting in the better zero-shot transfer. In contrast, the previous method can overfit to the unwanted tokens in the observation text that does not contribute to the decision making process.

	\vspace{-0.18cm}
	\section{Discussion}
	\vspace{-0.18cm}
	Empirical evaluation shows that our observation masking method can successfully reduce the overfitting problem in RL for Text-based games by reducing irrelevant tokens. Our method also learns at an accelerated rate requiring fewer training episodes due to pruned textual representations. We show that observation masking leads to better generalization, as demonstrated by superior performance for our CREST method with accelerated convergence with less number of training games as compared to the state-of-the-art method. 
	
	In this paper, we assume that the domain distribution between the training and evaluation are similar in our environments because our goal is to explore generalization by observation pruning without additional heuristic learning components. This means the evaluation games will have similar objectives as seen during the training games, and similar objects would be encountered in the evaluation games without encountering any novel objects. For example, if the goal objective is set as ``\textit{pickup the coin}'' in the training games, it will not be changed to ``\textit{eat the apple}'' which was never seen before in training. 
	
	To handle such environments with domain divergence, training needs to be performed with external datasets that show a satisfactory level of overlap with the domain of unseen test games. However, such training from external sources can be readily combined with our existing proposal in this paper using previous hand-crafted methods like \citet{adolphs2019ledeepchef, ammanabrolu2018playing}.

\end{document}

%% file: first_page_figure.tex
% !TEX root = main.tex

\begin{figure}[tb]
\begin{boxedminipage}{\columnwidth}
%\scriptsize
% \footnotesize
% \small
\textbf{Goal:}
Who's got a virtual machine and is about to play through an fast paced round of textworld? You do! Retrieve the coin in the balmy kitchen.
\newline \rule{\columnwidth}{1.2pt}
\textbf{Observation}:
You've entered a studio. \error{You try to gain information on your surroundings by using a technique you call ``looking.''} You need an unguarded exit ? you should try going \reph{east}. You need an unguarded exit? You should try \reph{going south}. \error{You don't like doors?} Why not try going \reph{west}, that entranceway is unblocked.
%\newline \textbf{Top relevant tokens:} {east,south,west,going,exit}
\newline% \rule{\columnwidth}{1.2pt} 
\textbf{Bootstrapped Policy Action:} \textit{go south}
%\newline \rule{\columnwidth}{0.4pt
%\textbf{Observation 2: Goal state}:
%You are in a kitchen. A balmy one. There is an unguarded exit to the \reph{north}. You need an unguarded exit? You should try going \reph{south}. There is a \reph{coin} on the floor .
%\newline \textbf{Action 2:} take coin
\end{boxedminipage}
\caption{Our method retains context-relevant tokens from the observation text~(shown in green) while pruning irrelevant tokens~(shown in red). A second policy network re-trained on the pruned observations generalizes better by avoiding overfitting to unwanted tokens. }
\label{fig_firstpage}
\end{figure}

%% file: main.bbl
\begin{thebibliography}{25}
\expandafter\ifx\csname natexlab\endcsname\relax\def\natexlab#1{#1}\fi

\bibitem[{Adhikari et~al.(2020)Adhikari, Yuan, C{\^o}t{\'e}, Zelinka, Rondeau,
  Laroche, Poupart, Tang, Trischler, and Hamilton}]{adhikari2020learning}
Ashutosh Adhikari, Xingdi Yuan, Marc-Alexandre C{\^o}t{\'e}, Mikul{\'a}{\v{s}}
  Zelinka, Marc-Antoine Rondeau, Romain Laroche, Pascal Poupart, Jian Tang,
  Adam Trischler, and William~L Hamilton. 2020.
\newblock Learning dynamic knowledge graphs to generalize on text-based games.
\newblock \emph{arXiv preprint arXiv:2002.09127}.

\bibitem[{Adolphs and Hofmann(2019)}]{adolphs2019ledeepchef}
Leonard Adolphs and Thomas Hofmann. 2019.
\newblock Ledeepchef: Deep reinforcement learning agent for families of
  text-based games.
\newblock \emph{arXiv preprint arXiv:1909.01646}.

\bibitem[{Ammanabrolu and Hausknecht(2020)}]{ammanabrolu2020graph}
Prithviraj Ammanabrolu and Matthew Hausknecht. 2020.
\newblock Graph constrained reinforcement learning for natural language action
  spaces.
\newblock \emph{arXiv preprint arXiv:2001.08837}.

\bibitem[{Ammanabrolu and Riedl(2018)}]{ammanabrolu2018playing}
Prithviraj Ammanabrolu and Mark~O Riedl. 2018.
\newblock Playing text-adventure games with graph-based deep reinforcement
  learning.
\newblock \emph{arXiv preprint arXiv:1812.01628}.

\bibitem[{Bahdanau et~al.(2014)Bahdanau, Cho, and Bengio}]{bahdanau2014neural}
Dzmitry Bahdanau, Kyunghyun Cho, and Yoshua Bengio. 2014.
\newblock Neural machine translation by jointly learning to align and
  translate.
\newblock \emph{arXiv preprint arXiv:1409.0473}.

\bibitem[{C{\^o}t{\'e} et~al.(2018)C{\^o}t{\'e}, K{\'a}d{\'a}r, Yuan, Kybartas,
  Barnes, Fine, Moore, Hausknecht, Asri, Adada et~al.}]{cote2018textworld}
Marc-Alexandre C{\^o}t{\'e}, {\'A}kos K{\'a}d{\'a}r, Xingdi Yuan, Ben Kybartas,
  Tavian Barnes, Emery Fine, James Moore, Matthew Hausknecht, Layla~El Asri,
  Mahmoud Adada, et~al. 2018.
\newblock Textworld: A learning environment for text-based games.
\newblock \emph{arXiv preprint arXiv:1806.11532}.

\bibitem[{Dhingra et~al.(2017)Dhingra, Li, Li, Gao, Chen, Ahmed, and
  Deng}]{dhingra2017towards}
Bhuwan Dhingra, Lihong Li, Xiujun Li, Jianfeng Gao, Yun-Nung Chen, Faisal
  Ahmed, and Li~Deng. 2017.
\newblock Towards end-to-end reinforcement learning of dialogue agents for
  information access.
\newblock In \emph{Proceedings of the 55th Annual Meeting of the Association
  for Computational Linguistics (Volume 1: Long Papers)}, pages 484--495.

\bibitem[{Fulda et~al.(2017)Fulda, Ricks, Murdoch, and Wingate}]{fulda2017can}
Nancy Fulda, Daniel Ricks, Ben Murdoch, and David Wingate. 2017.
\newblock What can you do with a rock? affordance extraction via word
  embeddings.
\newblock \emph{arXiv preprint arXiv:1703.03429}.

\bibitem[{He et~al.(2016)He, Chen, He, Gao, Li, Deng, and
  Ostendorf}]{he2016deep}
Ji~He, Jianshu Chen, Xiaodong He, Jianfeng Gao, Lihong Li, Li~Deng, and Mari
  Ostendorf. 2016.
\newblock Deep reinforcement learning with a natural language action space.
\newblock In \emph{Proceedings of the 54th Annual Meeting of the Association
  for Computational Linguistics (Volume 1: Long Papers)}, pages 1621--1630.

\bibitem[{Li et~al.(2017)Li, Chen, Li, Gao, and Celikyilmaz}]{li2017end}
Xiujun Li, Yun-Nung Chen, Lihong Li, Jianfeng Gao, and Asli Celikyilmaz. 2017.
\newblock End-to-end task-completion neural dialogue systems.
\newblock In \emph{Proceedings of the Eighth International Joint Conference on
  Natural Language Processing (Volume 1: Long Papers)}, pages 733--743.

\bibitem[{Liu and Singh(2004)}]{liu2004conceptnet}
Hugo Liu and Push Singh. 2004.
\newblock Conceptnet—a practical commonsense reasoning tool-kit.
\newblock \emph{BT technology journal}, 22(4):211--226.

\bibitem[{Mikolov et~al.(2013)Mikolov, Sutskever, Chen, Corrado, and
  Dean}]{mikolov2013distributed}
Tomas Mikolov, Ilya Sutskever, Kai Chen, Greg~S Corrado, and Jeff Dean. 2013.
\newblock Distributed representations of words and phrases and their
  compositionality.
\newblock In \emph{Advances in neural information processing systems}, pages
  3111--3119.

\bibitem[{Mnih et~al.(2015)Mnih, Kavukcuoglu, Silver, Rusu, Veness, Bellemare,
  Graves, Riedmiller, Fidjeland, Ostrovski et~al.}]{mnih2015human}
Volodymyr Mnih, Koray Kavukcuoglu, David Silver, Andrei~A Rusu, Joel Veness,
  Marc~G Bellemare, Alex Graves, Martin Riedmiller, Andreas~K Fidjeland, Georg
  Ostrovski, et~al. 2015.
\newblock Human-level control through deep reinforcement learning.
\newblock \emph{Nature}, 518(7540):529.

\bibitem[{Narasimhan et~al.(2015)Narasimhan, Kulkarni, and
  Barzilay}]{narasimhan2015language}
Karthik Narasimhan, Tejas Kulkarni, and Regina Barzilay. 2015.
\newblock Language understanding for text-based games using deep reinforcement
  learning.
\newblock In \emph{Proceedings of the 2015 Conference on Empirical Methods in
  Natural Language Processing}, pages 1--11.

\bibitem[{Osband et~al.(2016)Osband, Blundell, Pritzel, and
  Van~Roy}]{osband2016deep}
Ian Osband, Charles Blundell, Alexander Pritzel, and Benjamin Van~Roy. 2016.
\newblock Deep exploration via bootstrapped dqn.
\newblock In \emph{Advances in neural information processing systems}, pages
  4026--4034.

\bibitem[{Pathak et~al.(2017)Pathak, Agrawal, Efros, and
  Darrell}]{pathak2017curiosity}
Deepak Pathak, Pulkit Agrawal, Alexei~A Efros, and Trevor Darrell. 2017.
\newblock Curiosity-driven exploration by self-supervised prediction.
\newblock In \emph{Proceedings of the IEEE Conference on Computer Vision and
  Pattern Recognition Workshops}, pages 16--17.

\bibitem[{Pennington et~al.(2014)Pennington, Socher, and
  Manning}]{pennington2014glove}
Jeffrey Pennington, Richard Socher, and Christopher~D Manning. 2014.
\newblock Glove: Global vectors for word representation.
\newblock In \emph{Proceedings of the 2014 conference on empirical methods in
  natural language processing (EMNLP)}, pages 1532--1543.

\bibitem[{Plappert et~al.(2017)Plappert, Houthooft, Dhariwal, Sidor, Chen,
  Chen, Asfour, Abbeel, and Andrychowicz}]{plappert2017parameter}
Matthias Plappert, Rein Houthooft, Prafulla Dhariwal, Szymon Sidor, Richard~Y
  Chen, Xi~Chen, Tamim Asfour, Pieter Abbeel, and Marcin Andrychowicz. 2017.
\newblock Parameter space noise for exploration.
\newblock \emph{arXiv preprint arXiv:1706.01905}.

\bibitem[{Serban et~al.(2017)Serban, Sankar, Germain, Zhang, Lin, Subramanian,
  Kim, Pieper, Chandar, Ke et~al.}]{serban2017deep}
Iulian~V Serban, Chinnadhurai Sankar, Mathieu Germain, Saizheng Zhang, Zhouhan
  Lin, Sandeep Subramanian, Taesup Kim, Michael Pieper, Sarath Chandar,
  Nan~Rosemary Ke, et~al. 2017.
\newblock A deep reinforcement learning chatbot.
\newblock \emph{arXiv preprint arXiv:1709.02349}.

\bibitem[{Sutton et~al.(1998)Sutton, Barto et~al.}]{sutton1998introduction}
Richard~S Sutton, Andrew~G Barto, et~al. 1998.
\newblock \emph{Introduction to reinforcement learning}, volume~2.
\newblock MIT press Cambridge.

\bibitem[{Watkins and Dayan(1992)}]{watkins1992q}
Christopher~JCH Watkins and Peter Dayan. 1992.
\newblock Q-learning.
\newblock \emph{Machine learning}, 8(3-4):279--292.

\bibitem[{Wu et~al.(2016)Wu, Schuster, Chen, Le, Norouzi, Macherey, Krikun,
  Cao, Gao, Macherey et~al.}]{wu2016google}
Yonghui Wu, Mike Schuster, Zhifeng Chen, Quoc~V Le, Mohammad Norouzi, Wolfgang
  Macherey, Maxim Krikun, Yuan Cao, Qin Gao, Klaus Macherey, et~al. 2016.
\newblock Google's neural machine translation system: Bridging the gap between
  human and machine translation.
\newblock \emph{arXiv preprint arXiv:1609.08144}.

\bibitem[{Yin and May(2019)}]{yin2019learn}
Xusen Yin and Jonathan May. 2019.
\newblock Learn how to cook a new recipe in a new house: Using map
  familiarization, curriculum learning, and bandit feedback to learn families
  of text-based adventure games.
\newblock \emph{arXiv preprint arXiv:1908.04777}.

\bibitem[{Yuan et~al.(2018)Yuan, C{\^o}t{\'e}, Sordoni, Laroche, Combes,
  Hausknecht, and Trischler}]{yuan2018counting}
Xingdi Yuan, Marc-Alexandre C{\^o}t{\'e}, Alessandro Sordoni, Romain Laroche,
  Remi Tachet~des Combes, Matthew Hausknecht, and Adam Trischler. 2018.
\newblock Counting to explore and generalize in text-based games.
\newblock \emph{arXiv preprint arXiv:1806.11525}.

\bibitem[{Zahavy et~al.(2018)Zahavy, Haroush, Merlis, Mankowitz, and
  Mannor}]{zahavy2018learn}
Tom Zahavy, Matan Haroush, Nadav Merlis, Daniel~J Mankowitz, and Shie Mannor.
  2018.
\newblock Learn what not to learn: Action elimination with deep reinforcement
  learning.
\newblock In \emph{Advances in Neural Information Processing Systems}, pages
  3562--3573.

\end{thebibliography}
